\documentclass[10pt, conference]{ieeeconf}      % Use this line for a4 paper

\IEEEoverridecommandlockouts                              % This command is only needed if 
\usepackage{graphics} % for pdf, bitmapped graphics files
\usepackage{epsfig} % for postscript graphics files
\usepackage{mathptmx} % assumes new font selection scheme installed
\usepackage{times} % assumes new font selection scheme installed
\usepackage{amsmath} % assumes amsmath package installed
\usepackage{amssymb}  % assumes amsmath package installed
\usepackage{hyperref}  % assumes amsmath package installed
\usepackage{subfig}
\usepackage{xcolor}

\title{\LARGE \bf 
Navigation without localisation: reliable teach and repeat based on the convergence theorem
}

\author{Tom{\'a}{\v s} Krajn{\'i}k, Filip Majer, Lucie Halodov{\' a}, Tom{\'a}{\v s} Vintr
	\thanks{Artificial Intelligence Center, Faculty of Electrical Engineering, Czech Technical University {\tt\small{tomas.krajnik@fel.cvut.cz}}}
	\thanks{The work has been supported by projects 17-27006Y and CZ.02.1.01/0.0/0.0/16\_019/0000765.}
}

\begin{document}
\maketitle
\begin{abstract}
   We present a novel concept for teach-and-repeat visual navigation.
The proposed concept is based on a mathematical model, which indicates that in teach-and-repeat navigation scenarios, mobile robots do not need to perform explicit localisation.
Rather than that, a mobile robot which repeats a previously taught path can simply ``replay'' the learned velocities, while using its camera information only to correct its heading relative to the intended path. 
To support our claim, we establish a position error model of a robot, which traverses a taught path by only correcting its heading.
Then, we outline a mathematical proof which shows that this position error does not diverge over time. 
Based on the insights from the model, we present a simple monocular teach-and-repeat navigation method.
The method is computationally efficient, it does not require camera calibration, and it can learn and autonomously traverse arbitrarily-shaped paths.
In a series of experiments, we demonstrate that the method can reliably guide mobile robots in realistic indoor and outdoor conditions, and can cope with imperfect odometry, landmark deficiency, illumination variations and naturally-occurring environment changes. 
Furthermore, we provide the navigation system and the datasets gathered at \url{www.github.com/gestom/stroll_bearnav}.
		
\end{abstract}

\section{Introduction}

A considerable progress in visual-based systems capable of autonomous navigation of long routes was achieved during the last decade.
%The progress in this domain significantly contributed to the success of driverless vehicle projects like Google car or Tesla.
According to~\cite{survey,bonin2008visual}, vision-based navigation systems can be divided into map-less, map-based, and map-building based.
Map-less navigation systems such as~\cite{5594640,wang2014color,aastrand2005vision} aim to recognise traversable structures (e.g. roads, pathways, field rows, etc.) and use these to directly calculate motion commands.
Map-based navigation systems rely on environment models that are known apriori~\cite{kosaka1992fast}.
Map-building-based systems rely on maps for localisation and navigation as well, but they can build these maps themselves.
Some of these vision-based methods can build maps and localise the robot at the same time -- these are referred to as visual SLAM (Simultaneous Localisation and Mapping).

One of the most known visual SLAM systems, Monoslam~\cite{monoslam}, processes an image stream from an unconstrained-motion monocular camera in real-time, obtaining the trajectory of the camera and a 3D map of salient visual features~\cite{monoslam}.
Another method, the ORB-SLAM~\cite{orbslam,mur2017orb}, allows exploiting stereo and depth information to build both sparse and dense maps of the environment while estimating the camera motion in 6D.
Unlike the aforementioned systems, LSD-SLAM~\cite{lsdslam} and DSO~\cite{dso} do not rely on image feature extraction but create dense, large-scale maps by directly processing the intensities of the image pixels.
A recent, comprehensive review of SLAM systems is presented in~\cite{Cadena16tro-SLAMfuture}.
While being an important component of many navigation systems, SLAM by itself does not control the mobile robot motion, and thus it does not navigate robots per se.
Rather than that, it provides an environment map and a robot position estimate to the motion planning modules, which then guide the robot towards the desired goal.

Thus, one of the typical use of SLAM methods in practice is `teach-and-repeat', where a robot uses SLAM during a teleoperated drive, creating a map of the environment and use this map later on to repeat the taught path~\cite{stereo,teachrepeat,jakobysurfnav}.
This technique is analogous to a popular practice in industrial robotics, where an operator teaches a robot to perform some task simply by guiding its arm along the desired path.
The systems that use SLAM methods within the teach-and-repeat paradigm were extended by techniques like experience-based localisation~\cite{churchill}, feature selection~\cite{Dayoub11} or intrinsic image~\cite{finlaysonOntheremoval}, enabling their long-term deployment in environments that are challenging due to appearance changes~\cite{paton2016bridging,paton2015s,7989238} or difficult illumination conditions~\cite{paton2016dead}.

In the long-term deployments, it's assumed that the robots start and end their forays at their recharging stations with a known position, and occasional loss of localisation is solved by request for human intervention~\cite{strands}. 
Thus, teach-and-repeat methods employed in long-term scenarios do not typically address the kidnapped robot problem. 

Some of the teach-and-repeat systems do not rely on SLAM-built 3D maps of the environment, and employ visual servoing principles while respecting robot dynamics~\cite{kodit}.
For example~\cite{blanc2005indoor,matsumoto} create a visual path, which is a set of images along the human-guided route, and then employ visual servoing to guide robots across the locations these images were captured at.
Similarly,~\cite{segvic} represents the path as consecutive nodes, each containing a set of salient visual features, and uses local feature tracking to determine the robot steering to guide it to the next node.
The authors of~\cite{qualitative} extract salient features from the video feed on-the-fly and associate these with different segments of the teleoperated path.
When navigating a given segment, their robot moves forward and steers left or right based on the positions of the currently recognised and already mapped features.
The segment end is detected by means of comparing the mapped segment's last image with the current view.
The same navigation principle was recently deployed on micro aerial vehicles in~\cite{uav1}. 

At the time when the original SLAM-based teach-and-repeat framework was published~\cite{teachrepeat}, another article~\cite{jfr10} mathematically proved that (while being useful) explicit localisation is not necessary for teach-and-repeat scenarios.
The results of~\cite{jfr10} indicate that to repeat a taught path, camera input needs to be used only to correct the robot heading, leaving the position estimation to odometry.
The mathematical proof showed that for polygonal paths the heading corrections suppress odometric errors, preventing the overall position error of the robot to diverge.
The proof was supported by several long-term experiments~\cite{jfr10,pedre}, where a robot repeatedly traversed long paths in natural environments over the period of one year.
Thus, this SLAMless method showed good robustness to environment changes even without using experience-based or feature-preselection techniques~\cite{pedre}. 
However, the mathematical proof in~\cite{jfr10} was limited to paths consisting of straight segments.
Thus, the system based on~\cite{jfr10} could be taught only polygonal paths in a turn-move manner, and even a slight change of the movement direction during the teaching phase required to stop and turn the robot, which made the teaching tedious and deployment of the system rather impractical.

In this paper, we reformulate the problem presented in~\cite{jfr10} in a continuous rather than a discrete domain.
This allows us to simplify and extend the mathematical proof of~\cite{jfr10} to any continuous trajectory, not only polygonal paths as in~\cite{jfr10}.
A navigation system based on this extended formulation can thus be taught smooth, curved paths, which makes its teaching faster and navigation more efficient. 
 
The main contribution of this paper is the aforementioned mathematical proof which indicates that in teach-and-repeat scenarios, a robot can use its camera information only to correct its heading and it does not have to build metric maps or perform explicit localisation.
Based on this principle, we implement a teach-and-repeat navigation system and use it to verify the aforementioned hypothesis in realistic conditions. 
In a series of experiments, we compare the behaviour of the system with the proposed mathematical model and demonstrate the system's ability to reliably guide robots along the taught paths in adverse illumination conditions including night. 
Furthermore, we present the system as an open-source, ROS-based package and accompany the software with the datasets gathered during the experiments performed~\cite{stroll-bearnav}.
			
\section{Navigation Stability}\label{sec:convergence}

In this section, we analyse how heading correction influences the overall position error of a robot as it travels along a taught path.
At first, we establish a model of the robot movement along the desired path and we outline a model of the robot position error.
Then, we examine conditions under which the robot position error does not diverge.

\subsection{Paths with non-zero curvature}\label{sec:nonzero}

Let us assume that a human operator placed a robot at an initial position $x_p(0),y_p(0)$ and then she drove the robot by controlling its forward $v$ and angular $\omega$ velocities.
Let the robot record its $v$ and $\omega$ velocities together with the features detected in its camera image and let the robot index the features and velocities by the distance travelled.
Thus, the taught path $\mathcal{P}$ is defined by the initial point $x_p(0),y_p(0)$, velocity functions $v(d)$, and $\omega(d)$, where $d$ represents the length of the path from $x_p(0),y_p(0)$ to the current position.
Some locations (again indexed by $d$) on the trajectory are also associated with image coordinates and descriptors of the image features detected in the robot camera image.
Since the travelled distance $d$ depends on the robot forward velocity $v(d)$ by $d(t)=\int_0^t{v(d(\tau))d\tau}$, one can express the taught trajectory as $x_p(t)=x_p(d(t))$ and $y_p(t)=y_p(d(t))$.

Then, assume that a robot is placed at a position $x_{r}(0), y_{r}(0)$ and it is supposed to traverse the path $\mathcal{P}$.
The robot, having no information about its position or orientation, has to assume that it is at the start of the taught path, and thus its position estimate is $x_p(0),y_p(0)$. 
Therefore, the robot sets its forward and angular velocity to $v(0)$ and $\omega(0)$, respectively.
The robot also retrieves the image features it saw at $d=0$, matches them to the features in its current view and adjust its angular velocity in a way, which would decrease the horizontal distances (in the image coordinates) of the matched feature pairs.
As the robot moves forwards, it calculates the distance travelled $d$ and sets its forward velocity to $v(d)$ and angular velocity to $\omega(d) - \alpha\kappa$, where $\kappa$ is the most frequent horizontal displacement of the matched feature pairs and $\alpha$ is a user-set constant dependent on the robot dynamics and camera used. 
 
\begin{figure}[ht!]
\begin{center}
	\includegraphics[width=0.85\columnwidth]{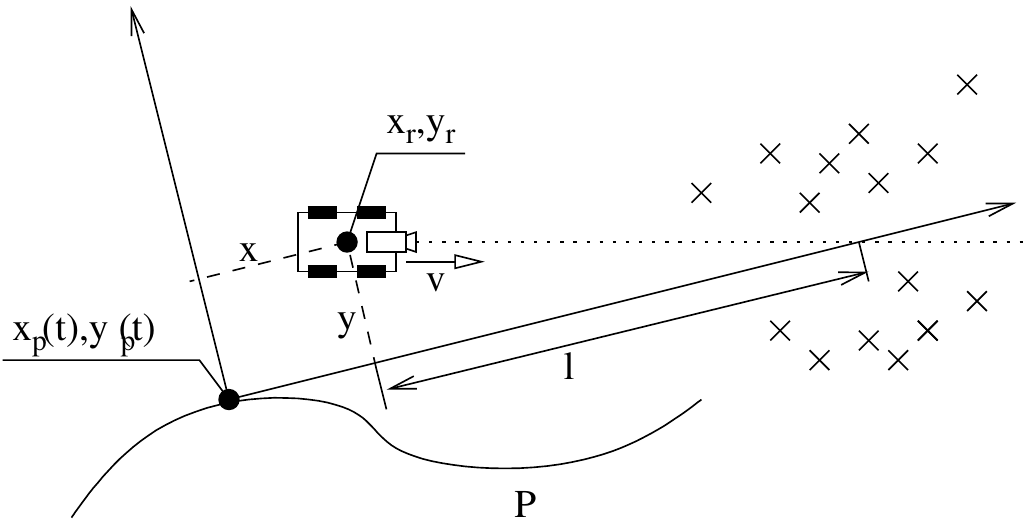}
	\caption{Robot position error chart. The robot at a position $x_r,y_r$ uses a local feature map of the taught path $\mathcal{P}$ at the position $x_p(t),y_p(t)$.}\label{fig:path}
\end{center}
\end{figure}
 
Since the robot camera is facing in the direction of the robot movement during the teaching phase, each feature that is in the current map lies in the vicinity of the tangent to the path at some finite distance, see Figure~\ref{fig:path}.
Assuming that the robot is able to turn fast enough to keep $\kappa$ low (i.e., it is able to turn so that the horizontal distances of the mapped/detected feature pairs are low), the direction of its current movement intersects the aforementioned tangent at a certain distance $l$.
The distance $l$ is related to the spatial distribution of the features in the traversed environment -- low $l$ is typical for cluttered environments and high $l$ occurs mostly in large open areas.

Figure~\ref{fig:path} illustrates that the error of the robot position estimate over time $x(t),y(t)$ can be defined as its position in a local coordinate frame defined by the location and orientation of the local map, which the robot uses to determine its velocities.
In other words, the robot position error $x(t),y(t)$ is defined by its position ($x_r(t),y_r(t)$) relatively to $x_p(t), y_p(t)$.
In order to analyse the evolution of the position error during the robot navigation, we form a differential equation describing ($\dot{x}=dx/dt,\dot{y}=dy/dt$):
\begin{equation}\label{eqn:movement}
\begin{array}{llllllllll}
\dot{x} &=& +&\omega\,{y}& -& v_r & +& v & +& s{_x}\\
\dot{y} &=& -&\omega\,{x}&&& - &v\,y\,l^{-1} & +& s{_y},
\end{array}
\end{equation}
where the terms $\omega\,{y}$, $\omega\,{x}$ and $-v_r$ are caused by the rotation and translation of the local coordinate system as the point $x_p(t),y_p(t)$ moves along the intended path, the term $+v$ is the movement of the robot along the path, $v\,y\,l^{-1}$ reflects the influence of the visual-based heading correction, and $s_x$ and $s_y$ represent perturbations caused by image processing imperfections, odometric errors and control system inaccuracies.
Since the errors $s_x$ and $s_y$ directly influence the velocities $\dot{x}$, $\dot{y}$ of the robot in our model, they encompass not only additive errors, such as occasional wheel slippage or imperfect image feature localisation, but also systematic errors, such as miscalibration of the odometry causing the  $v_r$ and $v$ to be different, or camera misalignment causing an offset in robot heading corrections.
Assuming that $v_r$ and $v$ are almost identical, and that their difference is included in the perturbance $s_x$, we can rewrite~(\ref{eqn:movement}) in a matrix form:
\begin{equation}\label{eqn:error}
\left(
\begin{array}{l}
\dot{x}\\\dot{y}
\end{array}
\right)
=
\left(
\begin{array}{rr}
0 & +\omega\\
-\omega &-v\,l^{-1}
\end{array}
\right)
\left(
\begin{array}{l}
x\\
y
\end{array}
\right)
+
\left(
\begin{array}{l}
s_x\\
s_y
\end{array}
\right),
\end{equation}
which is a system of linear differential equations.

In general, a linear continuous system $\mathbf{\dot{x}}=\mathbf{A}\mathbf{x}+\mathbf{s}$ is stable, i.e., the $\mathbf{x}$ does not diverge, if the real components of all eigenvalues of the matrix $\mathbf{A}$ are smaller than 0. 
Thus, the robot position error $x,y$ does not diverge if the matrix $\mathbf{A}$ from (\ref{eqn:error}) has all real components of its eigenvalues negative.
Eigenvalues of a 2$\times$2 matrix are obtainable by solving a quadratic equation
\begin{equation}\label{eqn:eigen}
\lambda\,(\lambda+v/l)+\omega^2 = 0, 
\end{equation}
\begin{equation}\label{eqn:eigen3}
\lambda_{1,2} = \frac{-{v/l}\pm\sqrt{(v/l)^2-4\omega^2}}{2}.
\end{equation}
In cases when $(v/l)^2 < 4\omega^2$, the expression $\sqrt{(v/l)^2 - 4\omega^2}$ is an imaginary number, and $\lambda_{1,2}$ are complex conjugates with the real component equal to $-v/(2l)$.
Since $v$ and $l$ are always positive, $-v/(2l)$ is always negative, which ensures the system stability and non-divergence of the robot position error $x,y$.
 
In the case of $(v/l)^2 \geq 4\omega^2$, the result of the square root is positive and the real part of the $\lambda_2$ eigenvalue  
\begin{equation}\label{eqn:eigen4}
\lambda_2 = \frac{-{v/l}-\sqrt{(v/l)^2-4\omega^2}}{2}
\end{equation}
is always negative.
The eigenvalue $\lambda_1$, which is calculated as  
\begin{equation}\label{eqn:eigen5}
\lambda_1 = \frac{-{v/l}+\sqrt{(v/l)^2-4\omega^2}}{2},
\end{equation}
is non-negative only if $\omega$ equals to 0.
\textit{This means, that if the robot does not move along a straight line, both eigenvalues of $\mathbf{A}$ are lower than 0, and therefore, both longitudinal ($x$) and lateral ($y$) components of the robot position error do not diverge even if the robot uses its exteroceptive sensors only to correct its heading.} $\square$

\subsection{Paths containing straight segments}

%However, the desired path might contain straight segments as well.
According to (\ref{eqn:movement}) and~(\ref{eqn:error}), if a robot travels along a straight line, its longitudinal position error, represented by $x$ in our model, gradually grows due to the perturbances $s_{x}$, which are caused primarily by odometric drift.
Let assume that the taught path consists of straight segments conjoined by segments with non-zero curvature.
Let assume that a robot started to traverse a straight segment at time $t_0$ and ended its traversal at $t_1$.
Its error after the segment traversal is obtainable by integrating~(\ref{eqn:error}) over time as: 
\begin{equation}\label{eqn:seg0}
\left(
\begin{array}{l}
x(t_1)\\y(t_1)
\end{array}
\right)
=
\left(
\begin{array}{rr}
1 & 0\\
0 & e^{-\frac{v}{l}(t_1-t_0)}
\end{array}
\right)
\left(
\begin{array}{l}
x(t_0)\\
y(t_0)
\end{array}
\right)
+
\left(
\begin{array}{l}
b_{x0}\\
b_{y0}
\end{array}
\right).
\end{equation}
Now, let assume that from the time $t_1$ to the time $t_2$, the robot traverses a segment with non-zero curvature.
Regardless of the segment shape and curvature, the magnitude of $x$ and $y$ decreases (see~(\ref{eqn:error})), and thus we can state that 
\begin{equation}\label{eqn:seg1}
\left(
\begin{array}{l}
x(t_2)\\y(t_2)
\end{array}
\right)
=
\mathbf{N_1}
\left(
\begin{array}{l}
x(t_1)\\
y(t_1)
\end{array}
\right)
+
\left(
\begin{array}{l}
b_{x1}\\
b_{y1}
\end{array}
\right),
\end{equation}
where the eigenvalues of the matrix $\mathbf{N_1}$ are lower than one.
Rewriting the discrete system (\ref{eqn:seg0}) in a compact form as $\mathbf{x_1}=\mathbf{N_0}\mathbf{x_0}+\mathbf{b_0}$, and (\ref{eqn:seg1}) as $\mathbf{x_2}=\mathbf{N_1}\mathbf{x_1}+\mathbf{b_1}$, and substituting (\ref{eqn:seg0}) into (\ref{eqn:seg1}) results in:
\begin{equation}\label{eqn:all}
\mathbf{x_2}
=
\mathbf{N_1}(\mathbf{N_0}\mathbf{x_0}+\mathbf{b_0})+\mathbf{b_1}=
\mathbf{N_1}\mathbf{N_0}\mathbf{x_0}+(\mathbf{N_1}\mathbf{b_0}+\mathbf{b_1}).
\end{equation}
Equation~\ref{eqn:all} represents a discrete system, which allows to estimate the robot position error after traversing a straight segment followed by a curved one. 
Since the largest eigenvalue of $\mathbf{N_0}$ equals to 1 (see (\ref{eqn:seg0})) and both eigenvalues of $\mathbf{N_1}$ are smaller than 1, the eigenvalues of their product $\mathbf{N_1N_0}$ are also smaller than 1.
This means that the discrete system (\ref{eqn:all}) is stable.
\textit{Thus, position error of a robot, which repeatedly traverses a path formed of conjoined straight and curved segments does not diverge even if it is using its exteroceptive sensors only to correct its heading.}$\square$

\subsection{Convergence proof assumptions}

The model that we established is based on two assumptions that might not be met in extreme cases.
First, it assumes that during the repeat phase, the robot perceives at least some image features that it saw during the teaching.
If the robot's initial position in the repeat phase is too far from the origin of the teaching step, or if the robot deviates from the path too much, the mapped features will not be in its field of view and the navigation will fail.
The actual position error that would cause the navigation to fail depends on robot's camera, path shape and feature distance.
In our experiments, we started the repeat phase with an initial position error exceeding 1~m, which is approximately an order of magnitude higher than the accuracy of the navigation, see Figures~\ref{fig:experiment2_1_error}, \ref{fig:experiment2_2_error}, and \ref{fig:experiment3_0_error}.
Thus, the typical navigation inaccuracy caused by $s_x$ and $s_y$ in (\ref{eqn:error}) is very unlikely to deviate from its path beyond the point where it can correct its position error.
In practice, a robot can monitor the consistency of the feature matching and when there are not enough correspondences, it can request human intervention.

The second assumption is that the robot steering controller gain (the parameter $\alpha$ in Section~\ref{sec:nonzero}) is set in a way, which allows the robot to align the mapped and currently visible features and steer itself as illustrated in Figure~\ref{fig:path}.
	
\section{Navigation method description}\label{sec:implementation}

The considered navigation system works in two steps: teach and repeat.
In the teaching phase, a robot is guided by an operator along a path, which is the robot supposed to autonomously navigate in the repeat phase. 
During the teaching, the robot extracts salient features from its onboard camera image and stores its current travelled distance and velocity.
During the autonomous navigation, the robot sets its velocity according to the travelled distance and compares the image coordinates of the currently detected and previously mapped features to correct its heading. 

\subsection{Image processing}

The feature extraction method which detects salient objects in the onboard camera image is a critical component of the navigation system because it is the only mechanism which the robot employs to reduce its position error.
Based on the results from our previous work on image feature stability in changing outdoor environments~\cite{grief}, we decided to use the Speeded Up Robust Features (SURF)~\cite{surf} and a combination of the AGAST~\cite{agast} and BRIEF~\cite{brief} methods.

The feature extraction is composed of two steps, detection of keypoints and description of their vicinity.
The keypoint detection indicates points in the image, which have sufficient contrast that makes them easy to localise and track.
In the case of SURF, the keypoint detection is based on the approximation of Hessian matrix determinant~\cite{surf}, while AGAST~\cite{agast} uses an optimised pixel brightness testing scheme around the keypoint candidate. 
To form the description of a particular keypoint, BRIEF~\cite{brief} calculates a binary descriptor by comparing brightnesses of randomly-chosen pixel pairs around the keypoint.
The advantage of this descriptor is an efficient calculation, low memory requirements, and rapid matching.
The SURF-based descriptor is based on image intensity gradients near the keypoint~\cite{surf}. 
While being slower to calculate and match, it is more resistant to large viewpoint changes.
%In our experiments, we use both SURF and AGAST/BRIEF features.

Once the keypoints are detected and described, they can be matched to the keypoints stored in a map and the associations can be used to correct the robot heading.
The quality of the features depends on the quality of the input image stream, which, in outdoor environments, suffers from varying illumination.
To compensate for the illumination instability, we select the exposure and brightness of the robot camera based on the results from~\cite{exposure,lucka_mesas}.

\subsection{Teaching (mapping) phase}\label{sec:map}

During the phase, the robot is driven through the environment by a human operator.
The robot continuously measures the distance it travelled and whenever the operator changes the forward or angular velocity, the robot saves the current distance and the updated velocity values -- we refer to this sequence as to a ``path profile''.
Additionally, the robot continuously extracts image features from its onboard camera image and every 0.2~m, it saves the currently detected image features in a local map, which is indexed by the current distance the robot travelled.

\subsection{Repeat (navigation) phase\label{sec:navi}}

At the start of this phase, the robot loads the path profile and creates a distance-indexed list of the local maps containing the image features.
Then, it sets its forward and angular velocities according to the first entry of the path profile and it loads the first local map containing data about image features visible at the start of the path.
As the robot moves forwards, it extracts image features from its onboard camera image and matches them to the ones loaded from the local map. 
The differences of the horizontal image coordinates of the matched feature pairs (i.e., the positions of the features in the camera image relative to the positions of the features in the preloaded map) are processed by a histogram voting method.
The maximum of the histogram indicates the most frequent difference in the horizontal positions of the features, which corresponds to the shift of the image that was acquired during the mapping phase relative to the image that is currently visible from the onboard camera.
This difference is used to calculate a corrective steering speed, which is added to the speed from the velocity profile.

If the histogram voting results are inconclusive due to the low number of features extracted, e.g., when the robot faces a featureless wall, the camera image is over- or under-exposed, etc., the corrective angular speed is not added to the one from the velocity profile.
In a case the visual information is not sufficient to determine the heading, the robot simply steers according to the path profile data. 
As the robot proceeds forwards along the taught path, it loads local maps and path profile data that correspond to the distance travelled and repeats the steps described above. 
Thus, the path profile allows the robot to steer approximately in the same way as during the teaching phase, and the image matching corrects the robot heading whenever it deviates from the intended path.

The principal advantage of the system is its robustness to feature deficiency and uneven feature distribution in the camera image. 
This makes the system robust to environment appearance changes and adverse lighting conditions.
The histogram voting-based heading corrections are demonstrated in videos available at~\cite{stroll-bearnav}.

\subsection{System implementation}

\begin{figure}
\begin{center}
	\includegraphics[width=0.85\columnwidth]{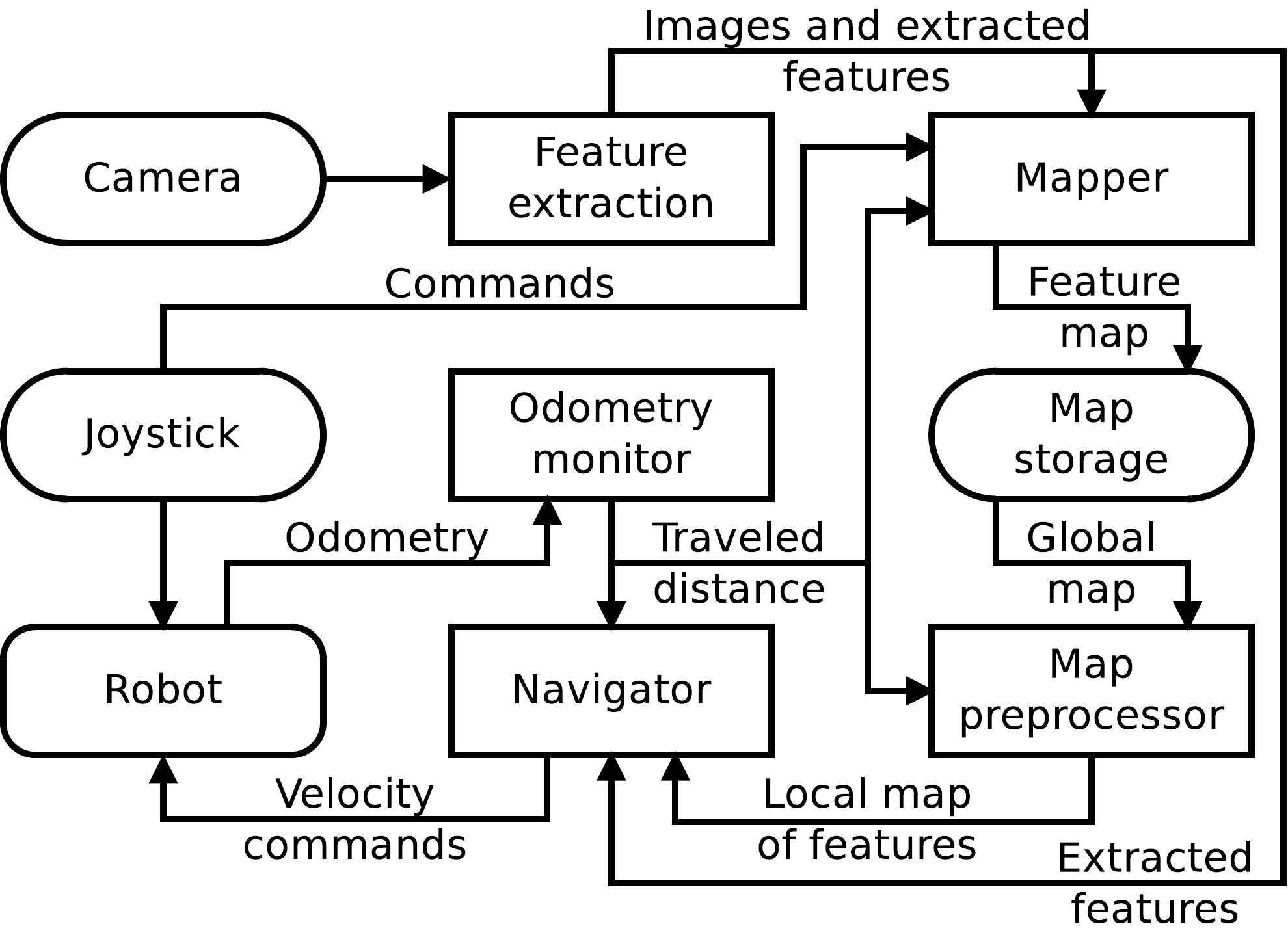}
	\caption{Software structure of the presented system\label{pic:system}}
\end{center}
\end{figure}

The navigation system was implemented in the Robotic Operating System (ROS), in particular the version Kinetic. 
The system structure is shown in Figure~\ref{pic:system}.
The \textit{feature extraction} node extracts image features from the robot camera and passes them to the \textit{mapper} and \textit{navigator} nodes.
The \textit{odometry monitor} node receives data from robot odometry and measures travelled distance. 
It also sends special messages every time the robot passes a given distance (e.g., 0.2~m) , which is used by the \textit{mapper node}, see Section~\ref{sec:map}.
The \textit{mapper} node receives features from the \textit{feature extraction} node and saves them into the local map when it receives the aforementioned message from the \textit{odometry monitor} node. It also saves the path profile. 
The \textit{map preprocessor} node loads all local maps and path profile, and then sends them to the \textit{navigator} node based on the travelled distance received from the \textit{odometry monitor}. 
The \textit{navigator} node receives the velocity profile and local feature maps and it matches the features from the maps to the currently visible features from the \textit{feature extraction} node.
It performs the histogram voting described in Section~\ref{sec:navi}, calculates the robot velocities and steers it along the path. 
All the aforementioned modules were implemented as ROS action servers with dynamically reconfigurable parameters and are available as C++ open source code at~\cite{stroll-bearnav}.
	
\section{Experimental evaluation}

The experimental evaluation of the proposed navigation system consists of 3 different experiments performed with 2 different robotic platforms.
The aim of the first experiment was to demonstrate that without the visual feedback or path profile information, the robot is not able to repeat the taught path.
The second experiment, which is performed under controlled conditions and an external localisation system, demonstrates the accuracy of the motion model established in Section~\ref{sec:convergence} and the ability of a minimalistic robotic platform to converge to the taught trajectory during multiple traversals of the intended path.
The third experiment demonstrates the trajectory convergence in challenging outdoor conditions including night.
%The fourth experiment demonstrates that the system is able to traverse long paths through a diverse environment.

\subsection{Platforms}

\begin{figure}[!ht]
\begin{center}
\hfill
\includegraphics[width=0.49\columnwidth]{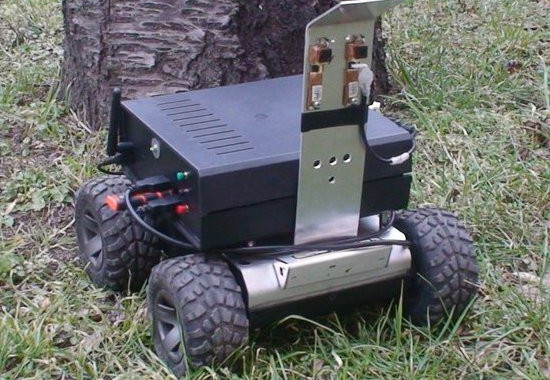}
\hfill
\includegraphics[width=0.49\columnwidth]{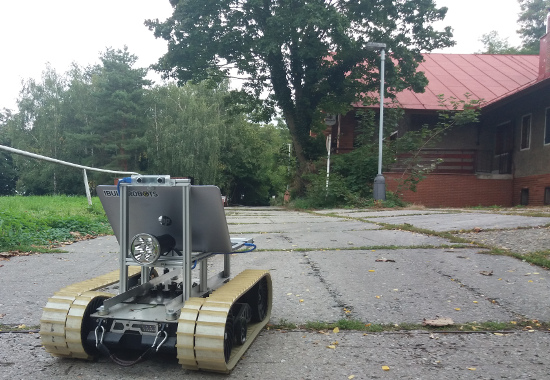}
\hfill
\caption{MMP-5 and Cameleon ECA used in our experiments.\label{pic:robots}}
\end{center}
\end{figure}

For indoor experiments, we used a lightweight robot based on an MMP-5 platform made by TheMachineLab~\ref{pic:robots}.
Its base dimensions are 0.3$\times$0.3$\times$0.1~m, its mass is 2.2~kg, and it can carry additional 2~kg payload while achieving speeds over 1.2~ms$^{-1}$.
The robot has a four-wheel differential drive with a control unit which allows setting PWM signal duty on individual motors by a simple serial protocol.
Since the platform does does not provide any odometric information, we estimate the travelled distance simply by the time and motors' PWM duty.
Its vision system is based on a single USB camera, which provides 320$\times$240 colour images with a 45$^\circ\times$35$^\circ$ field of view.
The computer is based on an AT3IONT-I miniATX board with an Intel Atom 330 CPU running at 1.6GHz with a 2GB RAM.
Due to its computational constraints, this robot is using the AGAST/BRIEF features.
 
For outdoor experiments, we used a heavy duty, tracked platform, called CAMELEON ECA, which is equipped with onboard PC and two cameras as primary sensors. 
Unfortunately, the cameras are positioned very low, which causes problems in grassy terrains and the onboard PC is too slow. 
Thus, we equipped the robot with a superstructure with several equipment mounts, where we placed the TARA USB stereo camera (we use only the left camera image in our experiments), Intel i3 laptop with 8GB RAM and the Fenix 4000 lumen torch, see Figure~\ref{pic:robots}.

\subsection{Experiment I: Proof-of-concept}

To evaluate the system's ability to repeat the taught path and to correct position errors that might arise during the navigation, we have taught the CAMELEON platform a closed, approximately 25~m long path in the outdoor environment in the Czech Technical University campus.
The shape of the trajectory is a closed, smooth, oval-shaped curve.
After mapping, we let the robot to drive along the taught path repeatedly.
Every time the robot completed a path loop, we measured its distance relative to the path end/start.
In this way, we quantitatively assess the robot's ability to adhere to the path it has been taught.
The experiments were performed during the evening, and therefore, the lighting conditions changed from high to low illumination, which made the image-based navigation particularly difficult.
Facing changes in environment and lighting conditions is inevitable for long-term navigation, that is why we chose this particular setup. 
 
To demonstrate the interplay between the path profile velocity setting and vision-based heading correction, we let the robot to drive the path autonomously using only the velocities remembered from the teaching phase (i.e. path profile), then only using visual information and then the combination of these. 
In the first test, we have deactivated the vision-based heading corrections and we let the robot move according to the path profile only -- this corresponds to the term $v\,y\,l^{-1}$ in (\ref{eqn:error}) being equal to 0.
The model (\ref{eqn:error}) predicts that the errors of $s_x$ and $s_y$ will gradually accumulate, drifting the robot off the taught path.
As predicted by the model, during this trial the robot slowly (by $\sim$1~m every loop) diverged from the taught path because of the inaccurate odometry. 

During the second trial, we have let the robot run with the method described in~\cite{jfr10}.  
Thus, the robot did not use the path profile information, but it moved forward with a constant speed and steered its heading according to the results of the image feature matching and histogram voting.
In this case, the robot diverged from the taught path as soon as it was supposed to perform a sharper turn, because the visual heading correction by itself could not perform sharp turns and the mapped features were lost from robot's field of view. 

The final trial used both path profile and visual feedback as described in Section~\ref{sec:implementation}. 
To verify if the robot can correct position errors that arise during navigation, we have started the autonomous navigation, not at the path start, but 0.9~m away in longitudal and 0.8~m away in lateral direction. 
The reason for the robot displacement is to demonstrate that unlike in the previous two cases, where the position error diverged, a combination of the vision and path profile information would allow the robot to suppress the error and adhere to the taught trajectory. 
As expected, each time the robot completed the taught path, its position error decreased, which confirms the assumptions stated in Section~\ref{sec:convergence}.
Further details about this experiment are provided in a short paper~\cite{bearnav_pair}.

\subsection{Experiment II: Position error model verification}

\begin{figure}[!ht]
\begin{center}
\includegraphics[width=0.95\columnwidth]{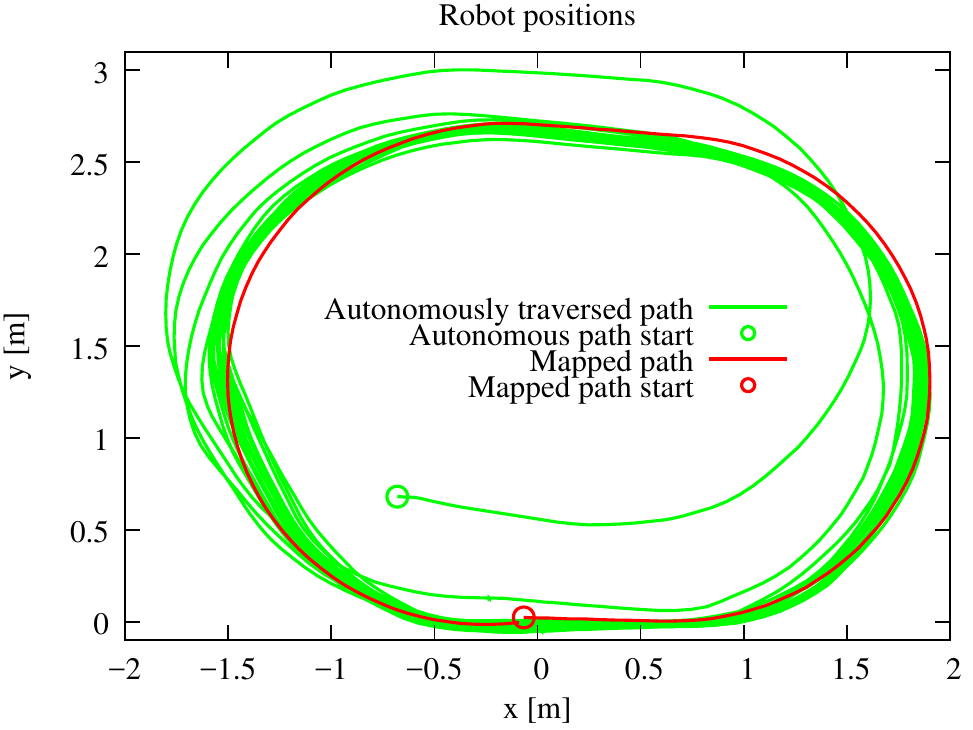}
\caption{Indoor trial I: Robot path during teach and repeat.\label{fig:experiment2_1}}
\end{center}
\end{figure}

The indoor experiments were meant to compare the real system behaviour with the model of the robot movement.
In these experimental trials, we used the MMP-5 robot platform equipped with a circular marker, which allows for an accurate tracking of the robot position.
In the first trial, we guided the robot along a 10~m long oval-shaped path consisting of two half-circle segments connected with two straight lines, see Figure~\ref{fig:experiment2_1}.
Then, we displaced the robot 1~m away from the path start and let it traverse along the path autonomously 20 times while recording its position with an external localisation system~\cite{whycon,cosphi}.

\begin{figure}[!ht]
\begin{center}
\includegraphics[width=0.95\columnwidth]{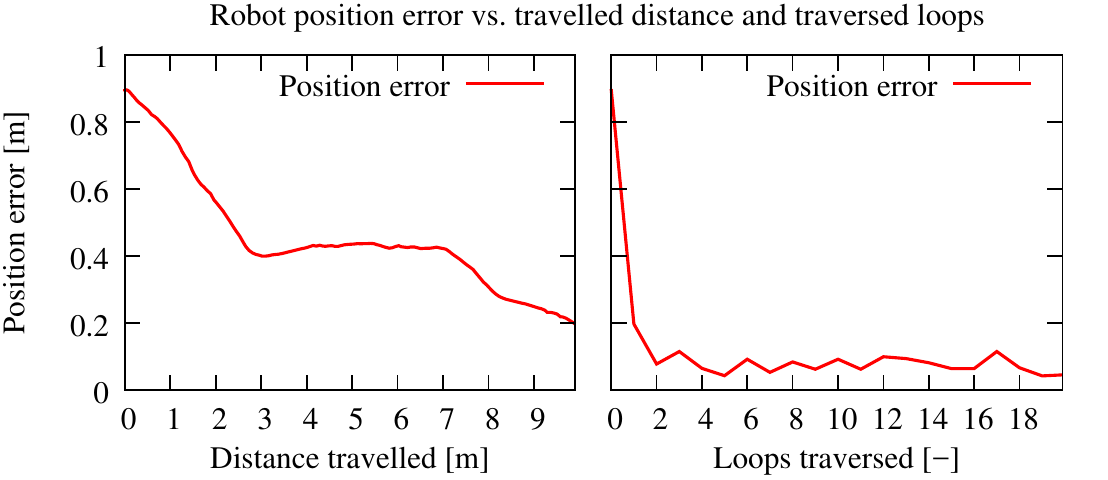}
\caption{Indoor trial I: Robot position error during the first autonomous path traversal (left) and during the 20 autonomous path repeats (right).\label{fig:experiment2_1_error}}
\end{center}
\end{figure}

Figure~\ref{fig:experiment2_1} shows the robot trajectory during the autonomous traversals, which slowly converges to the mapped path.
Each time the robot finished one path traversal, we measured its position relative to the path start.
Fig.~\ref{fig:experiment2_1_error} shows that the position error diminished after two loops, stabilising at 7~cm.

The convergence of the robot position during the autonomous travels is visible in Figure~\ref{fig:experiment2_1} and the robot position error evolution along the first path traversal in Figure~\ref{fig:experiment2_1_error}. 
The error evolution confirms the mathematical model presented in Section~\ref{sec:convergence} -- one can observe that during the first 5 meters, where the path is curved, the position error diminished rapidly.
Once the robot starts to traverse the straight segment, the position error stabilises and it starts to diminish once again when the robot starts to traverse the second semi-circular path segment.
Since the robot is nonholonomic, skid-steer drive, the convergence of the position error implies that its orientation conforms to the model~(\ref{eqn:movement}).

\begin{figure}[!ht]
\begin{center}
\includegraphics[width=0.95\columnwidth]{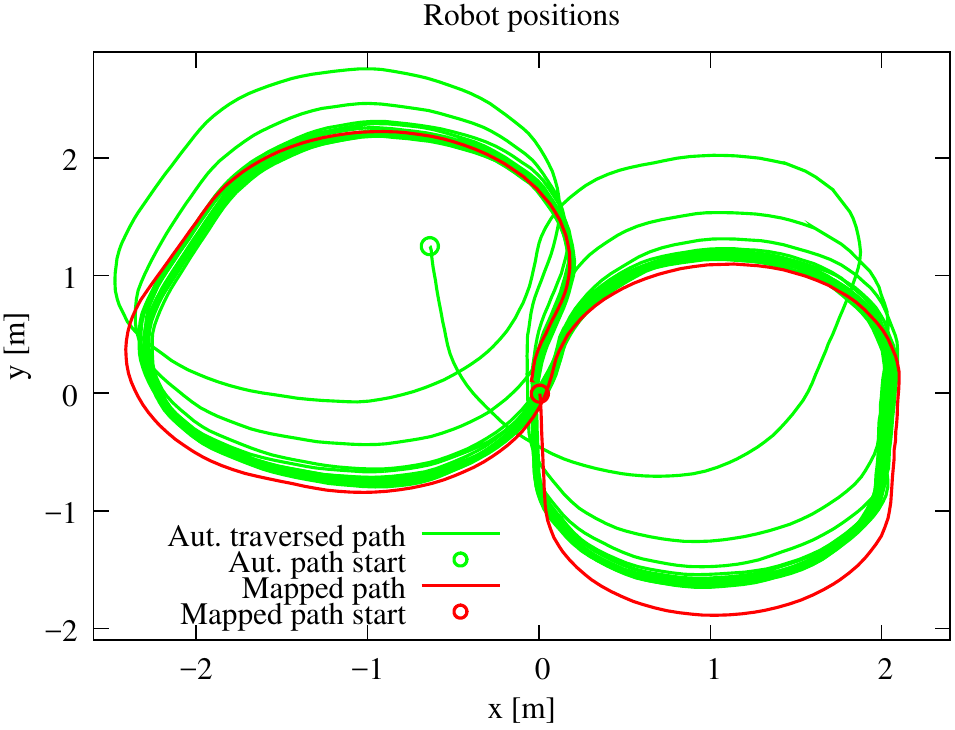}
\caption{Indoor trial II: Robot path during teach and repeat.\label{fig:experiment2_2}}
\end{center}
\end{figure}

\begin{figure}[!ht]
\begin{center}
\includegraphics[width=0.95\columnwidth]{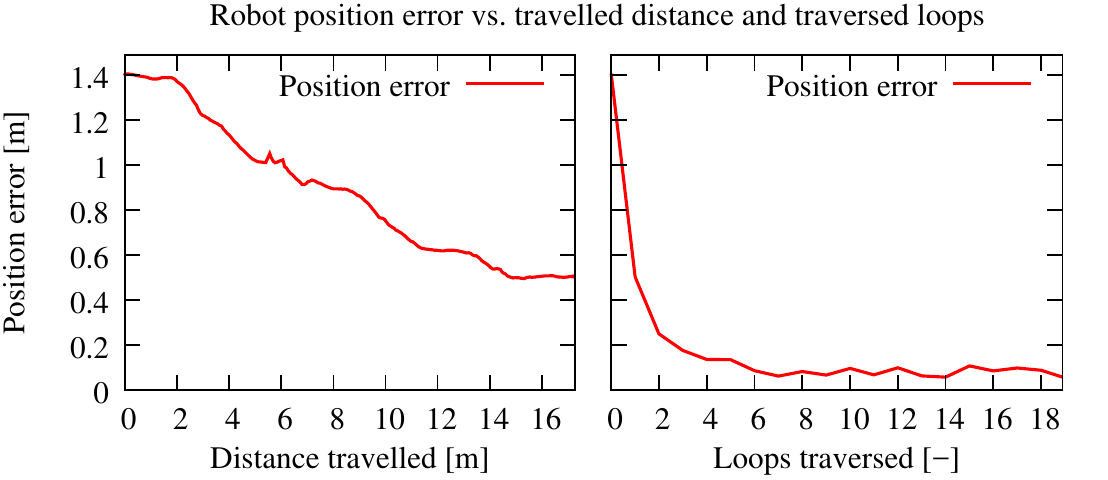}
\caption{Indoor trial II: Robot position error during the first autonomous path traversal (left) and during the 19 autonomous path repeats (right).\label{fig:experiment2_2_error}}
\end{center}
\end{figure}

In the second trial, we guided the robot along a 17~m long lemniscate-shaped path consisting of four half-circle segments connected with three straight ones and let it traverse the path 19 times, see Figure~\ref{fig:experiment2_2}.
This trial was performed in a larger hall than the previous one, and therefore, the average distance of landmarks is bigger than in the previous case, causing slower convergence of the robot trajectory, see Figure~\ref{fig:experiment2_2_error}.
Furthermore, the start and end of the taught path are about 0.15~m apart, which causes the robot to start with 0.15~m error during subsequent path traversals, which is notable in Figure~\ref{fig:experiment2_2}, where the first part of the repeated path is slightly displaced from the taught one.
Similarly to the previous case, the error evolution shown in the left part of Figure~\ref{fig:experiment2_2} conforms with the mathematical model presented in Section~\ref{sec:convergence} -- the error reduction is slower during traversal of the straight segments of the taught path.
In these experiments, a robot without any odometric system, equipped with a low resolution, uncalibrated camera which suffered from motion blur (see~\cite{stroll-bearnav}) reliably traversed over 700~m, reducing the initial $\sim$1~m position errors below $\sim$0.1~m.

\subsection{Experiment III: System robustness}

\begin{figure}[!ht]
\begin{center}
\includegraphics[width=0.45\columnwidth]{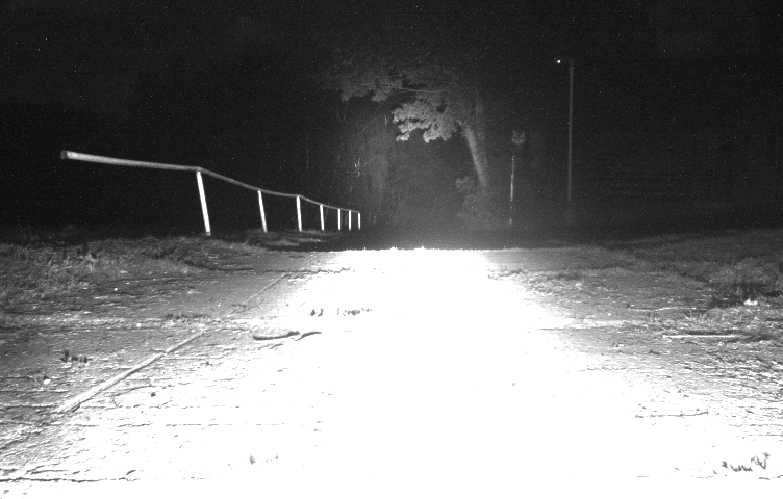}
\includegraphics[width=0.45\columnwidth]{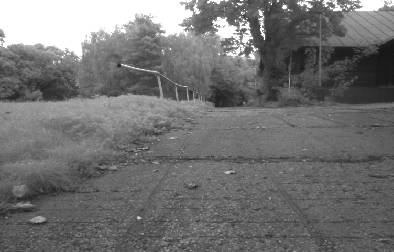}
\caption{Outdoor experiment: Hostibejk site at night and day from the robot perspective.\label{fig:daynight}}
\end{center}
\end{figure}

The purpose of final outdoor experiments is to demonstrate the ability of the system to cope with variable outdoor illumination. 
The first trial was performed during a day, where a robot was supposed to traverse a 60~m long path at the Hostibejk hill in Kralupy nad Vltavou, see Figures~\ref{pic:robots} and \ref{fig:daynight}.
The second and third trials were performed at night at the same location.
During the first trial, we created the map in a clear sky weather and let the robot traverse the path 7 times one month later during a cloudy day.

In the second trial, which was performed at night, we attached a 4000 lumen searchlight to the robot's superstructure.
The location of the trial was free of any artificial light sources; so, the most of the robot path, it saw only parts of the scene, which it illuminated itself, see Figure~\ref{fig:daynight}.
After creating the local map, we displaced the robot by 1.5~m and let it traverse the path 7 times (the number of traversals is limited by the capacity of the searchlight batteries). 
 
The third trial was performed at the same location, which at this time was partially illuminated, because three of the local street lamps were repaired in the meantime.
Again, after teaching the robot a 60~m long path, we displaced the robot by 1.5~m and let it traverse the path 7 times.
\begin{figure}[!ht]
\begin{center}
\includegraphics[width=0.95\columnwidth]{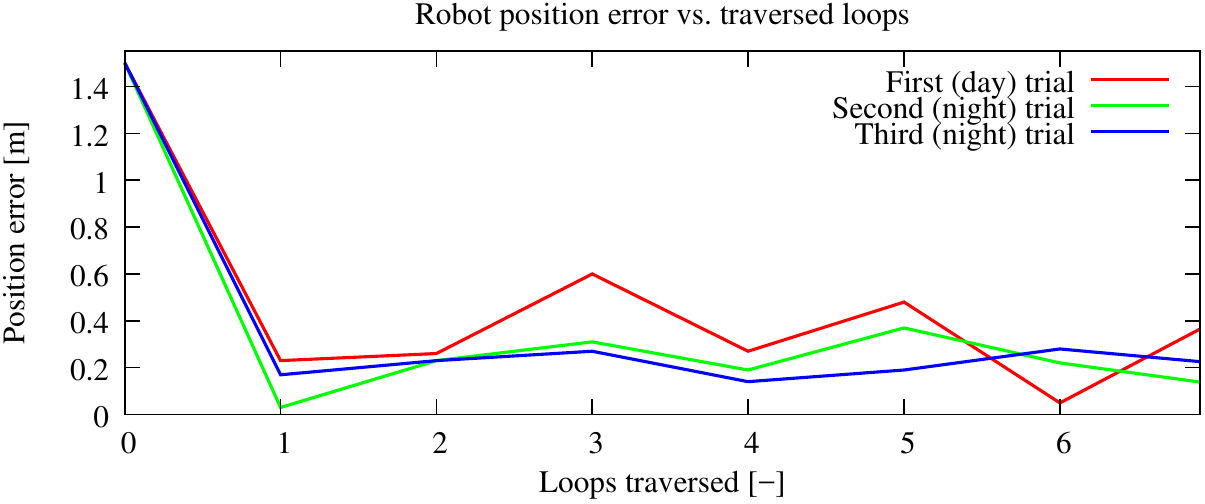}
\caption{Outdoor experiments: Robot position error relatively to the path start after traversing the path $n$ times.\label{fig:experiment3_0_error}}
\end{center}
\end{figure}
During these trials, we initiated the autonomous run 1.5~m from the taught path start and then measured the robot displacement from the path start every time it completed the taught path.
Figure~\ref{fig:experiment3_0_error} shows that the initial position error quickly diminished to values around 0.2--0.3~m.
In these experiments, a tracked robot with imperfect odometry, equipped with an uncalibrated camera working in low-visibility conditions (see~\cite{stroll-bearnav}) reliably traversed over 1.2~km, reducing the initial $\sim$1.5~m position errors to $\sim$0.2--0.3~m.
Furthermore, we did 10 days of additional tests at the same site over a period of one month.
In these tests, the robot successfully traversed the taught path 66 times, 21 during the day, 25 during the night and 20 during the sunset and covered distance over 4~km.

To compare the system's robustness to the state-of-the-art SLAM-based methods, we processed the data gathered during these trials by the ORB-SLAM~\cite{orbslam}, which we modified to take info account odometric information when initialising camera position estimation.  
To do so, we had to calibrate the camera and process the gathered data (in the form of ROSbags) by the ORB-SLAM. 
In our tests, we had to replay the ROSbags from the day trial 2.5$\times$ slower than in the original experiment several times because of occasional loss of tracking and subsequent breakdown of the ORB-SLAM method.
Despite trying several settings of ORB-SLAM, we could not build a consistent map from the night data due to feature deficiency and motion blur.
This comparison indicates the proposed method's robustness to difficult illumination conditions.
		
\section{Conclusion}

We formulated a mathematical proof which indicates that in teach-and-repeat scenarios, explicit localisation of a mobile robot along the taught route is not necessary.
Rather, a robot navigating through a known environment can use an environment map and visual feed to correct only its heading, while measuring the travelled distance only by its odometry.
Based on this principle, we designed and implemented a simple teach-and-repeat navigation system and evaluated its performance in a series of experimental trials. 
These experiments confirmed the validity of the aforementioned proof, showing that this kind of simple navigation is sufficient to keep the position error of the robot limited.
The requirement to establish only the robot heading simplifies the visual processing and makes it particularly robust to situations, where the detected and mapped features cannot be associated reliably.
This makes our system capable of handling difficult lighting conditions and environmental changes, which is demonstrated in several experiments.
Compared to the previous work~\cite{jfr10}, the mathematical proof of bearing-only navigation presented here is simpler, shorter and is not limited to polygonal routes only.
Thus, unlike in~\cite{jfr10}, where the robot could only learn polygonal routes in a turn move manner, our navigation system allows to learn arbitrarily-shaped routes, making its real-world deployment more feasible.
Furthermore, we show that the robot position error is asymptotically stable, whereas in our previous work~\cite{jfr10} we only proved its Lyapunov stability.
In our latest work, we extended the system so that it's able to improve its performance by exploiting the experience gathered during autonomous traversals~\cite{eliska_bp,lucka_mesas,lucka_bp}.

\bibliographystyle{IEEEtran}
\bibliography{main}

\end{document}